%% file: main.tex
\documentclass{article}

\usepackage{arxiv}

\usepackage[utf8]{inputenc} 
\usepackage[T1]{fontenc}    
\usepackage{hyperref}       
\usepackage{url}            
\usepackage{booktabs}       
\usepackage{amsfonts}       
\usepackage{nicefrac}       
\usepackage{microtype}      
\usepackage{lipsum}		
\usepackage{graphicx}
\usepackage{natbib}
\usepackage{doi}
\usepackage{booktabs}%
\usepackage{multirow}%
\usepackage{amsmath,amssymb,amsfonts}%
\usepackage{amsthm}%
\usepackage{mathrsfs}%
\usepackage{makecell}%
\usepackage{xcolor}%
\usepackage{pdflscape}%
\usepackage{algorithm}%
\usepackage[noend]{algpseudocode}%
\usepackage[capitalise]{cleveref}%
\usepackage{tikz}%
\usepackage{url}%
\usetikzlibrary{arrows.meta, calc, backgrounds, shadows, positioning, fit}

\definecolor{fg}{RGB}{13,122,80}
\definecolor{lg}{RGB}{234,246,240}
\definecolor{bk}{RGB}{51,51,51}
\definecolor{rg}{RGB}{190,215,200}
\definecolor{rb}{RGB}{190,200,215}
\definecolor{rlg}{RGB}{128,164,144}
\definecolor{rlb}{RGB}{128,144,164}

\title{Flow Matching-enabled Test-Time Refinement for Unsupervised Cardiac MR Registration}

\author{
	Yunguan Fu \\
	University College London \\
	InstaDeep \\
	\And
	Wenjia Bai \\
	Imperial College London \\
	\And
	Wen Yan \\
	University College London \\
	\And
	Matthew J Clarkson \\
	University College London \\
	\And
	Rhodri Huw Davies \\
	University College London \\
	\And
	Yipeng Hu \\
	University College London \\
}

\date{}

\renewcommand{\shorttitle}{Flow Matching-enabled Test-Time Refinement for Unsupervised Cardiac MR Registration}

\hypersetup{
pdftitle={A template for the arxiv style},
pdfsubject={q-bio.NC, q-bio.QM},
pdfauthor={David S.~Hippocampus, Elias D.~Striatum},
pdfkeywords={First keyword, Second keyword, More},
}

\begin{document}
\maketitle

\begin{abstract}
Diffusion-based unsupervised image registration has been explored for cardiac cine MR, but expensive multi-step inference limits practical use. We propose FlowReg, a flow-matching framework in displacement field space that achieves strong registration in as few as two steps and supports further refinement with more steps. FlowReg uses warmup-reflow training: a single-step network first acts as a teacher, then a student learns to refine from arbitrary intermediate states, removing the need for a pre-trained model as in existing methods. An Initial Guess strategy feeds back the model prediction as the next starting point, improving refinement from step two onward. On ACDC and MM2 across six tasks (including cross-dataset generalization), FlowReg outperforms the state of the art on five tasks (+0.6\% mean Dice score on average), with the largest gain in the left ventricle (+1.09\%), and reduces LVEF estimation error on all six tasks ($-2.58$ percentage points), using only 0.7\% extra parameters and no segmentation labels. Code is available at \url{https://github.com/mathpluscode/FlowReg}.
\end{abstract}


\section{Introduction}
Cardiovascular magnetic resonance (CMR) has been endorsed by clinical cardiology in international guidelines~\cite{rajiah2023cardiac} because it is non-invasive, radiation-free, and offers strong soft-tissue contrast. It is the reference standard for quantifying cardiac structure and function. As cardiovascular disease remains the leading global cause of death (about 19.8 million deaths per year~\cite{mensah2023heart}), CMR use continues to grow. Deformable image registration aligns CMR frames and supports motion tracking, strain quantification, and atlas construction. Classical methods such as ANTs~\cite{avants2008symmetric} are accurate but require minutes per pair, which is impractical clinically. Unsupervised learning methods~\cite{balakrishnan2019voxelmorph,chen2022transmorph,meng2024correlation} instead predict dense displacement fields (DDFs) in one forward pass, reducing inference to seconds.

Diffusion models~\cite{ho2020denoising} offer a generative alternative by iteratively denoising from a noisy start. This idea has been used in medical segmentation~\cite{wu2024medsegdiff,fu2023recycling} and registration. Tursynbek et al.~\cite{tursynbek2025guiding} use pre-trained diffusion features as similarity measures to guide registration networks. FSDiffReg~\cite{qin2023fsdiffreg} diffuses the fixed image and denoises a DDF from a pre-trained prediction. DRDM~\cite{zheng2026deformation} applies diffusion in deformation field space to synthesise diverse transformations for data augmentation and registration training. DiffuseReg~\cite{zhuo2024diffusereg} diffuses directly in DDF space conditioned on image pairs for pairwise registration, leveraging a pre-trained registration model. A key limitation of diffusion-based approaches is the large number of sampling steps needed at inference. Flow matching~\cite{lipman2022flow} addresses this with straighter interpolation paths that require fewer integration steps~\cite{liu2022flow}. Although it has been applied to image-to-image translation~\cite{disch2025temporal} and cardiac shape generation~\cite{ma2025cardiacflow}, it has not yet been applied to cardiac MR registration to the best of our knowledge.

We propose FlowReg, which formulates cardiac MR registration as iterative DDF refinement via flow matching. FlowReg surpasses the single-pass baseline in two steps and keeps improving with additional steps, without retraining and without the hundreds of samples often needed by diffusion methods. We introduce (1) warmup-reflow training to remove reliance on a pre-trained model, and (2) an Initial Guess strategy, inspired by recycling in diffusion-based segmentation~\cite{fu2023recycling}, that feeds back the model prediction as the next starting point. Evaluated on ACDC~\cite{bernard2018deep} and MM2~\cite{martin2023deep} for both end-diastole (ED) and end-systole (ES) target registration, including ACDC$\to$MM2 generalization, FlowReg consistently outperforms classical and learning-based baselines~\cite{avants2008symmetric,thirion1998image,balakrishnan2019voxelmorph,jia2023fourier,chen2022transmorph,shi2022xmorpher,chen2023transmatch,kebriti2025fractmorph,zhuo2024diffusereg,qin2023fsdiffreg} in both Dice score and clinical metrics (ejection fraction, myocardial thickness), with only 0.7\% extra parameters to CorrMLP~\cite{meng2024correlation}.

\section{Methods}
\begin{figure}[t]
\centering
\includegraphics[width=\linewidth]{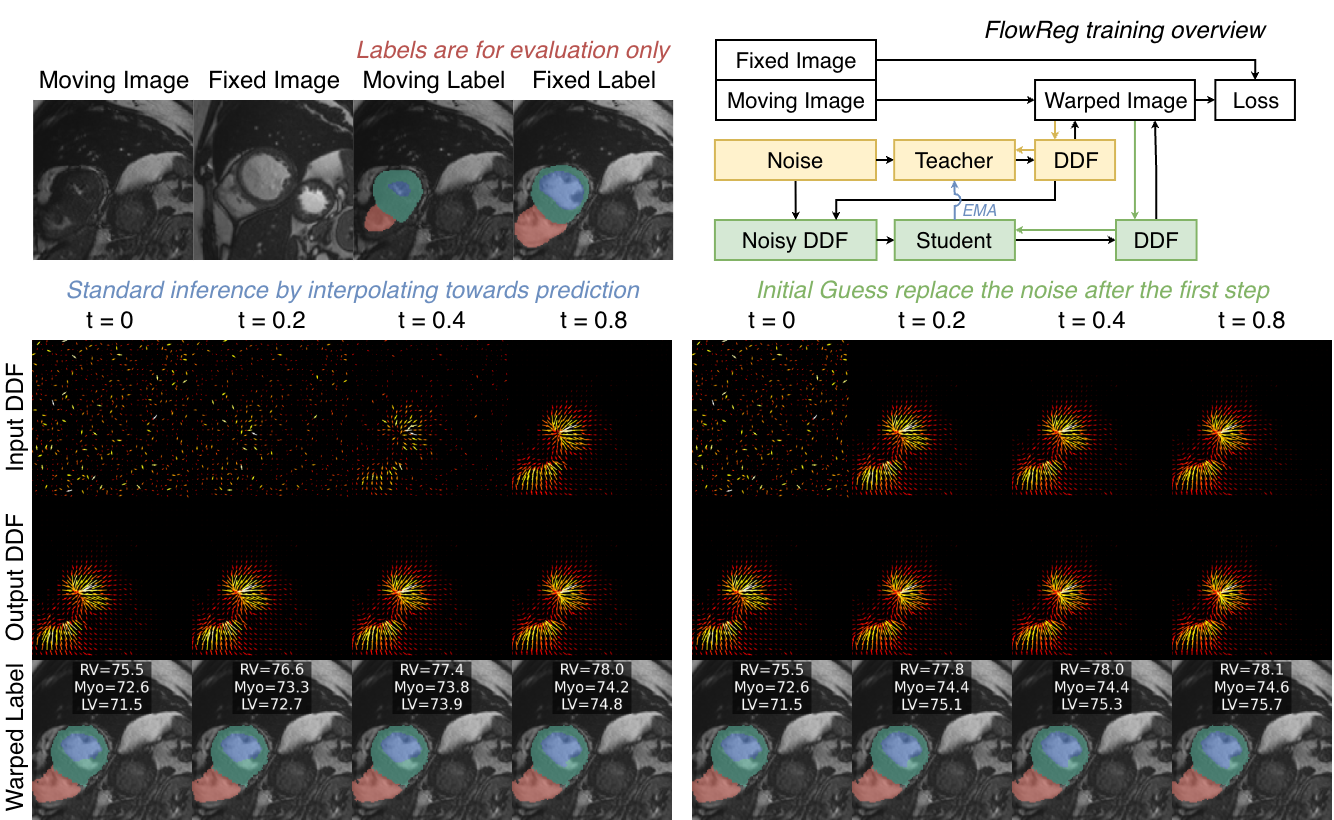}
\caption{Overview of FlowReg. Top-left: example image pair with labels (labels are used only for evaluation). Top-right: warmup-reflow training; the teacher predicts a reference DDF from noise, and the student learns from interpolated intermediate states. Yellow and green lines indicate the gradient propagation during warm-up and reflow training phases, respectively. Bottom-left: standard inference, which updates the DDF from pure noise ($t{=}0$) toward the prediction. Bottom-right: Initial Guess (IG), which replaces the noisy state after step one with the model prediction.}
\label{fig:overview}
\end{figure}

The registration task estimates a dense displacement field (DDF) $\psi\!: \mathbb{R}^3 \to \mathbb{R}^3$ for fixed image $I_f$ and moving image $I_m$, such that $(I_m\circ\psi)(\mathbf{x})=I_m(\psi(\mathbf{x}))\approx I_f$ for voxel $\mathbf{x}$, where $\circ$ denotes warping. Most deep registration methods learn $\psi=f_\theta(I_m, I_f)$. We instead model registration as a generative process $\psi=f_\theta(I_m, I_f, \psi_t, t)$, where the network also takes a noisy DDF $\psi_t$ and time $t$.

\textbf{Flow Matching.}
Flow matching~\cite{lipman2022flow} is a generative framework that learns to transport samples from a simple noise distribution to a target data distribution. It constructs straight-line interpolation paths $\psi_t = t \, \psi_1 + (1 - t) \, \varepsilon$ between noise $\varepsilon$ and data $\psi_1$, indexed by a continuous time $t\in[0,1]$. A neural network $f_\theta$ is trained to predict the velocity $v(\psi_t, t) = \frac{\hat{\psi}_1 - \psi_t}{1 - t}$ along these paths by minimizing an $\mathcal{L}^2$ loss between the predicted and target velocity. At inference, new samples are generated by integrating this learned velocity field from $t{=}0$ (noise) to $t{=}1$ (data), e.g.\ via Euler steps with step size $h$: $\psi_{t+h} = \psi_t + h \, v_\theta(\psi_t, t)$.

\textbf{Flow Matching in DDF Space.}
In our setting (\Cref{fig:overview}), $\psi_1$ is a DDF rather than an image, and the network is additionally conditioned on the image pair: $f_\theta(I_f, I_m, \psi_t, t)$. The noise $\varepsilon_i$ per voxel is sampled from $\mathcal{N}(0, \sigma_i^2)$ with $\sigma_i = 5 \min(\mathbf{s}) / s_i$ for $i=x,y,z$, where $\mathbf{s}=(s_x, s_y, s_z)$ is the voxel spacing.

\textbf{Warmup-Reflow Training.}
We use a registration loss $\mathcal{L}_\text{reg} = \mathcal{L}_\text{NCC} + \mathcal{L}_\text{grad}$, comprising a normalized cross-correlation loss ($\mathcal{L}_\text{NCC}$) on the warped image and an $\mathcal{L}^2$ gradient norm loss ($\mathcal{L}_\text{grad}$) on the predicted DDF. Both losses require no segmentation labels during training. However, a reference DDF is still needed for linear interpolation with noise. Instead of using a pre-trained model as in DiffuseReg~\cite{zhuo2024diffusereg}, we propose a warmup-reflow training scheme (\Cref{alg:train}).
\textit{Warmup}: A student model $f_\theta^\text{stu}$ is trained with only noise and $t=0$ for two epochs. This phase avoids a cold start by initializing the model as a single-step registration network; two epochs suffice, as the teacher continues to improve throughout reflow.
\textit{Reflow}: A teacher model $f_\theta^\text{tea}$ is initialized from the student and updated via exponential moving average (EMA). At each training step, a reference DDF $\psi^\text{tea}_1$ is obtained by $f_\theta^\text{tea}(I_f, I_m, \psi_0=\varepsilon, t=0)$. Training pair $(\psi_t, t)$ is then constructed by interpolation, where $t$ is drawn from a logit-normal distribution~\cite{esser2024scaling}. The student $f_\theta(I_f, I_m, \psi_t, t)$ is supervised with the same registration loss applied at arbitrary time steps. This phase trains the model to refine from intermediate states. We found that replacing the registration loss with the standard $\mathcal{L}^2$ velocity loss leads to unstable training, likely because no ground truth DDF is available and the teacher-generated targets are suboptimal.

\input{algo}

\begin{figure}[t]
\centering
\scalebox{0.75}{\input{network}}
\hfill
\scalebox{0.75}{\input{corrmlp_network}}
\hfill
\scalebox{0.75}{\input{flowreg_network}}
\caption{Network architecture. Left: shared encoder extracts four-scale feature pairs $(F_m^k, F_f^k)$ from images $(I_m, I_f)$; the decoder predicts DDFs $\psi^k$ coarse-to-fine ($k{=}4$ coarsest, $k{=}1$ finest). Centre: CorrMLP decoder stages 4 and 3. Right: FlowReg decoder stages 4 and 3. W: warp. $\uparrow$: upsample. $+$: addition. Green elements are the FlowReg additions: timestep conditioning ($t$), a noisy DDF input ($\psi_t^4$) used to warp moving features at the coarsest stage, and residual correction.}
\label{fig:arch}
\end{figure}

\textbf{Architecture.}
FlowReg follows CorrMLP~\cite{meng2024correlation} (\Cref{fig:arch}) with encoder-decoder design. The timestep $t$ is encoded by sinusoidal embedding and added to moving features. A downsampled noisy DDF $\psi_t^4$ is used to warp moving features and predict a residual correction to $\psi_t^4$ at the coarsest stage. Other stages follow CorrMLP. Overall, FlowReg adds only timestep conditioning, increasing parameters by 28K parameters, about 0.7\% over CorrMLP (4.2M).

\textbf{Inference.}
We follow EDM~\cite{karras2022elucidating} to use Heun's second-order method to add stochastic noise to convert the ODE to an SDE. At each step (\Cref{alg:inference}), we move from the current DDF $\psi_t$ towards the predicted DDF $\hat{\psi}_1$ with the estimated velocity. We also add guidance~\cite{bansal2023universal} so that each inference step further reduces $\mathcal{L}_\text{reg}$; this can be viewed as a single-step instance optimisation~\cite{balakrishnan2019voxelmorph} applied within each sampling step.

\textbf{Initial Guess.}
Empirically, we observed that the first prediction from pure noise is not very accurate, which is common in flow matching and diffusion models~\cite{geng2025mean}. We therefore propose an Initial Guess (IG) method: after the first inference step at time $t$, instead of taking a one-step interpolation between $\psi_{t_0}{=}\psi_0{=}\varepsilon$ and $\hat{\psi}_1{=}f_\theta(I_f,I_m,\varepsilon,0)$ (which remains noisy), we set $\psi_{t_1}=\hat{\psi}_1$ so that later steps can refine this better starting point.

\section{Experiment Settings}
We evaluate on ACDC~\cite{bernard2018deep} (90/10/50 train/val/test) and MM2~\cite{martin2023deep} (156/38/157 train/val/test). Each subject has ED and ES frames with left ventricle (LV), myocardium (Myo), and right ventricle (RV) labels, used only for evaluation. Volumes are resampled to $1.5 {\times} 1.5 {\times} 3.15$\,mm, center-cropped to $128{\times}128{\times}32$, and augmented with random flips and $90^\circ$ rotations, following FSDiffReg~\cite{qin2023fsdiffreg}. We define six tasks: ED$\to$ES and ES$\to$ED on both datasets, plus cross-dataset generalization (trained on ACDC, tested on MM2) in both directions. Here $A{\to}B$ denotes registering moving frame $A$ to fixed frame $B$.
We report Dice score (RV, Myo, LV, mean, foreground), deformation regularity ($\%|J_\psi| {\leq} 0$, $\mathrm{STD}(|J_\psi|)$), and mean absolute error (MAE) for left and right ventricle ejection fraction (LVEF, RVEF) and myocardial thickness (MT). EF compares true and warped moving-frame segmentations; MT compares true and warped fixed-frame segmentations.
FlowReg was trained with Adam (learning rate\,$= 10^{-4}$), batch size 1, for 100 epochs with early stopping (patience 10). FSDiffReg~\cite{qin2023fsdiffreg} and CorrMLP~\cite{meng2024correlation} were retrained from open-source code under identical data conditions. Inference hyperparameters were tuned on ACDC ES$\to$ED and fixed for all tasks. For noise injection, we set $\eta$ to saturate churn $\gamma=\min(\eta/N,\sqrt{2}-1)$. We swept $\lambda_g\in\{0.01,0.02,0.05,0.1\}$; performance was stable for $\lambda_g\le0.05$, while $\lambda_g{=}0.1$ reduced accuracy by $0.1\%$. Unless noted, FlowReg uses $N{=}10$ with Initial Guess and $\mathcal{L}_\text{reg}$ guidance ($\lambda_g{=}0.05$). For instance optimisation, we use Adam with learning rate 0.01 (selected from $\{0.1,0.01,0.001\}$).

\section{Results}
We first register ES to ED on ACDC (\Cref{tab:baseline}). Among existing approaches, CorrMLP is the strongest method (83.88\% mean Dice score). While one-step FlowReg (82.94\%) is lower, two steps (84.27\%) surpass CorrMLP, and 10 steps reach 84.68\%. This comes with slightly higher deformation irregularity (0.44\% negative Jacobian at 10 steps vs.\ 0.24\% for CorrMLP), though folding remains low. The gains generalize across tasks and datasets (\Cref{tab:dsc}): FlowReg has the best mean Dice score, with Bonferroni-corrected significance on five of six tasks ($p < 0.05$, paired t-test), averaging 80.44\% vs.\ 79.81\% (CorrMLP). The largest gain is for LV (+1.09\%). In cross-dataset tests, FlowReg still leads, suggesting good generalization across scanners and populations.
\input{table_baseline}
\input{table_dsc}

We next study multi-step behavior (\Cref{tab:iterative}). For cascaded CorrMLP, we repeatedly warp the moving image with the predicted DDF and re-register it to the fixed image. CorrMLP peaks at 2 steps (79.83\%) and then drops to 79.37\% at 20 steps, likely because increasingly aligned inputs drift from the training distribution of unregistered pairs. FlowReg instead improves with more steps, reaching 80.44\% at 10 steps before plateauing. Unlike CorrMLP, FlowReg refines the DDF directly along the learned ODE trajectory.
\input{table_iterative}
\input{table_clinical}

\Cref{tab:iterative} ablates inference components. Heun-only ODE integration reaches 80.24\% at 10 steps; adding SDE noise raises this to 80.38\%, suggesting better exploration around suboptimal trajectories. SDE with Euler alone gives 80.33\%; combining SDE and Heun gives 80.38\%, indicating complementary effects: Heun reduces discretization error, SDE broadens exploration. Initial Guess gives the largest low-step gain (2-step: 79.76\%$\to$80.14\%): without it, early steps recover from noisy initialization; with it, refinement starts from the model prediction. Guidance gives small but consistent gains (e.g.\ +0.06\% at 10 steps) by directly improving image similarity at each denoising step. Gains diminish at higher step counts as trajectories converge. We also apply instance optimisation to the outputs of CorrMLP and FlowReg (\Cref{tab:iterative}). Although instance optimisation yields marginal gains, two-step FlowReg remains better than CorrMLP, showing that learned multi-step refinement is additive to test-time optimisation.

To test whether Dice score gains translate to clinical utility, we report ejection fraction and myocardium thickness errors (\Cref{tab:clinical}). FlowReg reduces LVEF and RVEF error on all six tasks (averaging 12.73\% vs.\ 15.31\% and 9.79\% vs.\ 11.44\%, all $p < 0.05$). This is consistent with the higher LV/RV Dice scores as more accurate ventricular volumes lead to more accurate EF. We also observe larger EF errors when registering to ES than to ED, consistent with lower Dice scores on ES targets and the higher sensitivity of smaller end-systolic cavities to registration error. Myocardial thickness error is similar on average (0.81 vs.\ 0.82\,mm). Overall, multi-step refinement improves estimation of clinically relevant cardiac biomarkers.

\section{Discussion}
Formulating cardiac MR registration as iterative DDF refinement via flow matching yields a different test-time behavior from single-pass models: accuracy improves with more inference steps, without retraining. Warmup-reflow enables training from scratch, and Initial Guess lets FlowReg surpass the single-pass baseline in two steps. Across six tasks on two datasets, FlowReg consistently outperforms baselines in both Dice and clinical metrics. Several limitations deserve discussion. First, one-step FlowReg is weaker than CorrMLP; this is consistent with the known difficulty of single-step generation in flow matching, and the gains come from multi-step refinement. Techniques such as mean flows~\cite{geng2025mean} that improve single-step quality could potentially close this gap. Second, deformation irregularity increases with step count, though absolute folding rates stay low; future work should examine where folding occurs and whether diffeomorphic constraints can reduce it without harming accuracy. Third, we observe high consistency between predictions across seeds empirically (Dice score $> 95\%$), and leveraging stochastic sampling for diverse predictions and uncertainty estimation is a promising direction. While the EF and myocardium thickness results confirm clinical relevance, whether these gains extend to other downstream tasks, such as myocardial strain quantification, remains to be validated. Future work will explore improved single-step quality, diffeomorphic extensions, and applying FlowReg to broader medical image registration tasks.

\section{Acknowledgement}
This research has been conducted using the UK Biobank Resource under Application Number 71702. The authors acknowledge the use of resources provided by the Isambard-AI National AI Research Resource (AIRR). Isambard-AI is operated by the University of Bristol and is funded by the UK Government’s Department for Science, Innovation and Technology (DSIT) via UK Research and Innovation; and the Science and Technology Facilities Council [ST/AIRR/I-A-I/1023]. RD is directly and indirectly supported by the NIHR Biomedical Research Centres at University College London Hospital and Barts Health NHS Trusts. WB acknowledge the support of EPSRC CVD-Net Grant (EP/Z531297/1 ) and BHF New Horizon Grant (NH/F/23/70013).

\bibliographystyle{unsrtnat}
\bibliography{references}

\end{document}

%% file: algo.tex
\begin{figure}[t]
\begin{minipage}[t]{0.45\textwidth}
\begin{algorithm}[H]
\caption{FlowReg Training}
\label{alg:train}
\small
\begin{algorithmic}[1]
\Require $I_f$, $I_m$, epoch $m$, $\mu=0.99$
\State $\varepsilon_i{\sim}\mathcal{N}(0, \sigma_i^2), i{=}x,y,z$
\If{$m \leq 2$} \Comment{Warmup}
    \State $\hat{\psi}_1{\gets}f_\theta(I_f,\, I_m,\, \varepsilon,\, 0)$
\Else \Comment{Reflow}
    \State $\psi_1^\text{tea}{\gets}f_\theta^\text{tea}(I_f,\, I_m,\, \varepsilon,\, 0)$
    \State $t{\gets}\mathrm{sigmoid}(z),\; z \sim \mathcal{N}(0, 1)$ 
    \State $\psi_t{\gets}t\psi_1^\text{tea} + (1 - t)\varepsilon$ 
    \State $\hat{\psi}_1{\gets}f_\theta(I_f,\, I_m,\, \psi_t,\, t)$ 
\EndIf
\State $\mathcal{L}{\gets}\mathcal{L}_{\text{NCC}}(I_m{\circ}\hat{\psi}_1,\, I_f) + \mathcal{L}_{\text{grad}}(\hat{\psi}_1)$
\State Update $\theta$ via $\nabla_\theta \mathcal{L}$
\State $\theta^\text{tea}{\gets}\mu \, \theta^\text{tea} + (1 - \mu) \, \theta$ 
\end{algorithmic}
\end{algorithm}
\end{minipage}
\begin{minipage}[t]{0.50\textwidth}
\begin{algorithm}[H]
\caption{FlowReg Inference}
\label{alg:inference}
\small
\begin{algorithmic}[1]
\Require $I_f$, $I_m$, steps $N$, $\eta$, $\lambda_g$
\Function{FWD}{$\psi, t$}{$\rightarrow\hat{\psi}_1$} \Comment{Guidance}
    \State $\hat{\psi}_1{\gets}f_\theta(I_f,\, I_m,\, \psi,\, t)$
    \State $\hat{\psi}_1{\gets}\hat{\psi}_1 - \lambda_g \nabla_{\hat{\psi}_1}[\mathcal{L}_{\text{NCC}} {+} \mathcal{L}_{\text{grad}}]$
\EndFunction
\State $\varepsilon_i{\sim}\mathcal{N}(0, \sigma_i^2),i{=}x,y,z$; $\psi{\gets}\varepsilon$; $h{\gets}1/N$
\For{$i = 0, \ldots, N{-}1$}
    \State $t{\gets}i/N$; $\sigma{\gets}1{-}t$;
    \State $\hat{\sigma}{\gets}\sigma(1{+}\min(\eta/N,\, \sqrt{2}{-}1))$
    \State $\psi{\gets}\psi + \sqrt{\hat{\sigma}^2 - \sigma^2}\,\varepsilon$ if $i > 0$ \Comment{SDE}
    \State $\hat{\psi}_1{\gets}\Call{FWD}{\psi, t}$
    \State $v_1{\gets}(\hat{\psi}_1 - \psi) / (1 - t)$
    \If{$i<N-1$} \Comment{Heun}
        \State $\tilde{\psi}{\gets}\psi + v_1 \cdot h$; $\tilde{\psi}_1{\gets}\Call{FWD}{\tilde{\psi}, t_{i+1}}$
        \State $v_2{\gets}(\tilde{\psi}_1 - \tilde{\psi}) / (1 - t_{i+1})$
        \State $v_1{\gets}(v_1 + v_2) / 2$
    \EndIf
    \State $\psi{\gets}\hat{\psi}_1$ if $i=0$ else $\psi + v_1 \cdot h$ \Comment{IG}
\EndFor
\end{algorithmic}
\end{algorithm}
\end{minipage}
\end{figure}

%% file: network.tex
\begin{tikzpicture}[
    >={Stealth[length=4pt, width=3pt]},
    ka/.style  = {->, bk, line width=0.7pt},
    kl/.style  = {bk, line width=0.7pt},
    ga/.style  = {->, fg, line width=0.7pt},
    stage/.style={draw=bk, line width=0.7pt, rectangle,
                  minimum width=1.6cm, minimum height=0.7cm,
                  align=center, rounded corners=1.5pt, fill=white,
                  font=\footnotesize},
]
\def\yone{-5.5}
\def\ytwo{-4}
\def\ythree{-2.5}
\def\yfour{-1.0}
\def\encX{0}
\def\tX{1.5}
\def\featX{2.8}
\def\decX{5.0}
\def\psiX{6.7}

\node[stage] (e1) at (\encX, \yone) {Stage 1};
\node[stage] (e2) at (\encX, \ytwo) {Stage 2};
\node[stage] (e3) at (\encX, \ythree) {Stage 3};
\node[stage] (e4) at (\encX, \yfour) {Stage 4};
\node[draw=bk, line width=0.7pt, rounded corners=3pt, inner xsep=8pt, inner ysep=8pt,
      fit=(e1)(e4), label={[anchor=south]above:Encoder}] (encbox) {};
\node (input) at (\encX, \yone - 1.5) {$(I_m,\, I_f)$};
\draw[ka] (input) -- (e1.south);
\draw[ka] (e1) -- (e2);
\draw[ka] (e2) -- (e3);
\draw[ka] (e3) -- (e4);

\node (F4) at (\featX, \yfour) {$(F_m^4,\, F_f^4)$};
\node (F3) at (\featX, \ythree) {$(F_m^3,\, F_f^3)$};
\node (F2) at (\featX, \ytwo) {$(F_m^2,\, F_f^2)$};
\node (F1) at (\featX, \yone) {$(F_m^1,\, F_f^1)$};

\draw[ka] (e1.east) -- (F1.west);
\draw[ka] (e2.east) -- (F2.west);
\draw[ka] (e3.east) -- (F3.west);
\draw[ka] (e4.east) -- (F4.west);

\node[fg] (tnode) at (\tX, 0.5) {$t$};
\draw[ga] (tnode.south) -- (\tX, \yfour);
\draw[ga] (\tX, \yfour) -- (\tX, \ythree);
\draw[ga] (\tX, \ythree) -- (\tX, \ytwo);
\draw[ga] (\tX, \ytwo) -- (\tX, \yone);

\node[stage] (s4) at (\decX, \yfour) {Stage 4};
\node[stage] (s3) at (\decX, \ythree) {Stage 3};
\node[stage] (s2) at (\decX, \ytwo) {Stage 2};
\node[stage] (s1) at (\decX, \yone) {Stage 1};
\node[draw=bk, line width=0.7pt, rounded corners=3pt, inner xsep=8pt, inner ysep=8pt,
      fit=(s1)(s4), label={[anchor=north]below:Decoder}] (decbox) {};
\draw[ka] (F4.east) -- (s4.west);
\draw[ka] (F3.east) -- (s3.west);
\draw[ka] (F2.east) -- (s2.west);
\draw[ka] (F1.east) -- (s1.west);
\node[fg] (psit) at (\decX, 0.5) {$\boldsymbol{\psi}_t^{\,4}$};
\draw[ga] (psit) -- (s4.north);
\node (psi4) at (\psiX, {0.5*(\yfour+\ythree)}) {$\boldsymbol{\psi}^4$};
\node (psi3) at (\psiX, {0.5*(\ythree+\ytwo)}) {$\boldsymbol{\psi}^3$};
\node (psi2) at (\psiX, {0.5*(\ytwo+\yone)}) {$\boldsymbol{\psi}^2$};
\node (psi1) at (\psiX, \yone) {$\boldsymbol{\psi}^1$};
\draw[ka] (s4.east) -| (psi4);
\draw[ka] (psi4.west) -- (\decX, {0.5*(\yfour+\ythree)}) -- (s3.north);
\draw[ka] (s3.east) -| (psi3);
\draw[ka] (psi3.west) -- (\decX, {0.5*(\ythree+\ytwo)}) -- (s2.north);
\draw[ka] (s2.east) -| (psi2);
\draw[ka] (psi2.west) -- (\decX, {0.5*(\ytwo+\yone)}) -- (s1.north);
\draw[ka] (s1.east) -- (psi1);
\end{tikzpicture}

%% file: corrmlp_network.tex
\begin{tikzpicture}[
    >={Stealth[length=4pt, width=3pt]},
    ka/.style  = {->, bk, line width=0.7pt},
    kl/.style  = {bk, line width=0.7pt},
    kd/.style  = {bk, line width=0.7pt, dashed},
    kad/.style = {->, bk, line width=0.7pt, dashed},
    addk/.style  = {circle, draw=bk, line width=0.7pt, minimum size=13pt,
                    inner sep=0pt, fill=white, font=\small, text=bk},
    warpk/.style = {circle, draw=bk, line width=0.7pt, minimum size=15pt,
                    inner sep=0pt, fill=white,
                    font=\scriptsize, text=bk},
    blk/.style   = {draw=bk, line width=0.7pt,
                    minimum width=2.6cm, minimum height=0.65cm,
                    rounded corners=1.5pt, fill=white,
                    font=\footnotesize},
]

\def\Ax{0.7}      
\def\Bx{1.7}      
\def\Cx{1.2}     
\def\Ux{2.8}      
\def\Dx{4.1}      
\def\Fx{4.6}      
\def\Ex{5.1}      

\def\rF{0}         
\def\rW{-0.9}      
\def\rCa{-1.9}     
\def\rDash{-3.3}   
\def\rCb{-3.3}     
\def\rBot{-4.3}    
\def\rPb{-5.2}     

\node[bk, font=\normalsize\itshape] (Fm4) at (\Ax, \rF) {$F_m^4$};
\node[bk, font=\normalsize\itshape] (Ff4) at (\Bx, \rF) {$F_f^4$};

\node[blk] (C4) at (\Cx, \rCa) {CorrMLP Block};
\draw[ka] (Fm4.south) -- (Fm4.south |- C4.north);
\draw[ka] (Ff4.south) -- (Ff4.south |- C4.north);

\pgfmathsetmacro{\dashY}{(\rCa + \rCb) / 2}
\draw[kl] (C4.south) -- (\Cx, \dashY);
\draw[ka] (\Cx, \dashY) -- (\Cx, \rBot+0.15);

\node[bk, font=\normalsize\itshape] (psi4) at (\Cx, \rBot) {$\psi^4$};

\node[blk] (C3b) at (\Fx, \rCb) {CorrMLP Block};
\path (C3b.north west) ++(0.8, 0) coordinate (c3btarget);
\draw[kd] (\Cx, \dashY) -- (c3btarget |- 0, \dashY);
\draw[kad] (c3btarget |- 0, \dashY) -- (c3btarget);

\node[bk, font=\normalsize\itshape] (Fm3) at (\Dx, \rF) {$F_m^3$};
\node[bk, font=\normalsize\itshape] (Ff3) at (\Ex, \rF) {$F_f^3$};

\node[warpk] (W3) at (\Dx, \rW) {W};
\draw[ka] (Fm3) -- (W3);

\node[blk] (C3a) at (\Fx, \rCa) {CorrMLP Block};
\draw[ka] (W3.south) -- (W3.south |- C3a.north);
\draw[ka] (Ff3.south) -- (Ff3.south |- C3a.north);

\path (C3b.north east) ++(-0.8, 0) coordinate (c3br);
\draw[kl] (C3a.south) -- (\Fx, \dashY);
\draw[kl] (\Fx, \dashY) -- (c3br |- 0, \dashY);
\draw[ka] (c3br |- 0, \dashY) -- (c3br);

\node[addk] (res3) at (\Fx, \rBot) {$+$};
\draw[ka] (C3b.south) -- (res3);

\node[bk, font=\normalsize\itshape] (psi3) at (\Fx, \rPb) {$\psi^3$};
\draw[ka] (res3) -- (psi3);

\node[addk] (up) at (\Ux, \rBot) {$\uparrow$};
\draw[ka] (psi4) -- (up);

\draw[ka] (up) -- (res3);

\draw[kl] (up.north) -- (\Ux, \rW);
\draw[ka] (\Ux, \rW) -- (W3.west);

\end{tikzpicture}

%% file: flowreg_network.tex
\begin{tikzpicture}[
    >={Stealth[length=4pt, width=3pt]},
    ga/.style  = {->, fg, line width=0.7pt},
    gl/.style  = {fg, line width=0.7pt},
    ka/.style  = {->, bk, line width=0.7pt},
    kl/.style  = {bk, line width=0.7pt},
    kd/.style  = {bk, line width=0.7pt, dashed},
    kad/.style = {->, bk, line width=0.7pt, dashed},
    addg/.style  = {circle, draw=fg, line width=0.7pt, minimum size=13pt,
                    inner sep=0pt, fill=white, font=\small, text=fg},
    addk/.style  = {circle, draw=bk, line width=0.7pt, minimum size=13pt,
                    inner sep=0pt, fill=white, font=\small, text=bk},
    warpg/.style = {circle, draw=fg, line width=0.7pt, minimum size=15pt,
                    inner sep=0pt, fill=white,
                    font=\scriptsize, text=fg},
    warpk/.style = {circle, draw=bk, line width=0.7pt, minimum size=15pt,
                    inner sep=0pt, fill=white,
                    font=\scriptsize, text=bk},
    blk/.style   = {draw=bk, line width=0.7pt,
                    minimum width=2.6cm, minimum height=0.65cm,
                    rounded corners=1.5pt, fill=white,
                    font=\footnotesize},
]

\def\Px{-0.4}     
\def\Ax{0.7}      
\def\Bx{1.7}      
\def\Cx{1.2}     
\def\Mx{2.0}      
\def\Ux{2.8}      
\def\Dx{4.1}      
\def\Fx{4.6}      
\def\Ex{5.1}      

\def\rT{0}
\def\rA{-0.85}
\def\rF{-1.7}
\def\rW{-2.55}
\def\rCa{-3.5}
\def\rDash{-4.9}
\def\rCb{-4.9}
\def\rBot{-5.9}
\def\rPb{-6.75}

\node[fg, font=\normalsize\itshape] (t) at (\Ax, \rT) {$t$};
\node[addg] (t4) at (\Ax, \rA) {$+$};
\draw[ga] (t) -- (t4);
\node[addg] (t3) at (\Dx, \rA) {$+$};
\draw[gl] (t.east) -- ++(.2,0) -- (\Dx, \rT);
\draw[ga] (\Dx, \rT) -- (t3);

\node[fg, font=\normalsize\itshape] (Fm4) at (\Ax, \rF) {$F_m^4$};
\draw[ga] (t4) -- (Fm4);
\node[bk, font=\normalsize\itshape] (Ff4) at (\Bx, \rF) {$F_f^4$};
\node[warpg] (W4) at (\Ax, \rW) {W};
\draw[ga] (Fm4) -- (W4);
\node[fg, font=\normalsize\itshape] (pt4) at (\Px, \rW) {$\psi_t^4$};
\draw[ga] (pt4) -- (W4);

\node[blk] (C4) at (\Cx, \rCa) {CorrMLP Block};
\draw[ka] (W4.south) -- (W4.south |- C4.north);
\draw[ka] (Ff4.south) -- (Ff4.south |- C4.north);

\pgfmathsetmacro{\dashY}{(\rCa + \rCb) / 2}
\draw[kl] (C4.south) -- (\Cx, \dashY);
\node[addg] (res4) at (\Cx, \rBot) {$+$};
\draw[ka] (\Cx, \dashY) -- (res4.north);

\node[blk] (C3b) at (\Fx, \rCb) {CorrMLP Block};
\path (C3b.north west) ++(0.8, 0) coordinate (c3btarget);
\draw[kd] (\Cx, \dashY) -- (c3btarget |- 0, \dashY);
\draw[kad] (c3btarget |- 0, \dashY) -- (c3btarget);

\draw[gl] (pt4.south) -- (\Px, \rBot);
\draw[ga] (\Px, \rBot) -- (res4.west);

\node[fg, font=\normalsize\itshape] (psi4) at (\Mx, \rBot) {$\psi^4$};
\draw[ga] (res4) -- (psi4);

\node[bk, font=\normalsize\itshape] (Fm3) at (\Dx, \rF) {$F_m^3$};
\draw[ga] (t3) -- (Fm3);
\node[bk, font=\normalsize\itshape] (Ff3) at (\Ex, \rF) {$F_f^3$};
\node[warpk] (W3) at (\Dx, \rW) {W};
\draw[ka] (Fm3) -- (W3);

\node[blk] (C3a) at (\Fx, \rCa) {CorrMLP Block};

\draw[ka] (W3.south) -- (W3.south |- C3a.north);

\draw[ka] (Ff3.south) -- (Ff3.south |- C3a.north);

\path (C3b.north east) ++(-0.8, 0) coordinate (c3br);
\draw[kl] (C3a.south) -- (\Fx, \dashY);              
\draw[kl] (\Fx, \dashY) -- (c3br |- 0, \dashY);      
\draw[ka] (c3br |- 0, \dashY) -- (c3br);              

\node[addk] (res3) at (\Fx, \rBot) {$+$};
\draw[ka] (C3b.south) -- (res3);

\node[bk, font=\normalsize\itshape] (psi3) at (\Fx, \rPb) {$\psi^3$};
\draw[ka] (res3) -- (psi3);

\node[addk] (up) at (\Ux, \rBot) {$\uparrow$};
\draw[ka] (psi4) -- (up);

\draw[ka] (up) -- (res3);

\draw[kl] (up.north) -- (\Ux, \rW);
\draw[ka] (\Ux, \rW) -- (W3.west);

\end{tikzpicture}

%% file: table_baseline.tex
\begin{table}[!b]
\centering
\fontsize{8}{9.6}\selectfont
\caption{Dice score on ACDC ES$\to$ED. Mean: average Dice score of RV, Myo, and LV. Foreground: Dice score on merged RV+Myo+LV. $\%|J_\psi|\leq0$ and $\mathrm{STD}(|J_\psi|)$ measure deformation regularity (lower is better). Time: inference seconds per sample on one GPU. \textbf{Bold} indicates the best value per column.}
\label{tab:baseline}
\begin{tabular}{llllll}
\toprule
Method & Mean & Foreground & $\%|J_\psi|\leq 0$ & $\text{STD}(|J_\psi|)$ & Time (s) \\
\midrule
ANTs\cite{avants2008symmetric} & 65.98\textsubscript{$\pm$12.25} & 83.70\textsubscript{$\pm$6.84} & 0.93\textsubscript{$\pm$3.01} & 0.08\textsubscript{$\pm$0.03} &  \\
Demons\cite{thirion1998image} & 71.23\textsubscript{$\pm$9.73} & 83.43\textsubscript{$\pm$6.16} & 0.77\textsubscript{$\pm$0.42} & 0.40\textsubscript{$\pm$0.07} &  \\
VoxelMorph\cite{balakrishnan2019voxelmorph} & 74.88\textsubscript{$\pm$8.69} & 86.08\textsubscript{$\pm$5.99} & 0.02\textsubscript{$\pm$0.03} & 0.14\textsubscript{$\pm$0.03} &  \\
Fourier-Net\cite{jia2023fourier} & 60.68\textsubscript{$\pm$12.09} & 83.09\textsubscript{$\pm$6.01} & \textbf{0.00}\textsubscript{$\pm$0.00} & \textbf{0.03}\textsubscript{$\pm$0.01} &  \\
TransMorph\cite{chen2022transmorph} & 68.11\textsubscript{$\pm$11.70} & 82.87\textsubscript{$\pm$7.22} & 0.06\textsubscript{$\pm$0.01} & 0.15\textsubscript{$\pm$0.05} &  \\
XMorpher\cite{shi2022xmorpher} & 70.09\textsubscript{$\pm$9.53} & 84.96\textsubscript{$\pm$4.04} & 0.02\textsubscript{$\pm$0.04} & 0.13\textsubscript{$\pm$0.03} &  \\
TransMatch\cite{chen2023transmatch} & 47.25\textsubscript{$\pm$8.54} & 86.19\textsubscript{$\pm$4.76} & 0.03\textsubscript{$\pm$0.04} & 0.14\textsubscript{$\pm$0.03} &  \\
FractMorph\cite{kebriti2025fractmorph} & 75.15\textsubscript{$\pm$8.95} & 86.45\textsubscript{$\pm$4.72} & 0.05\textsubscript{$\pm$0.04} & 0.15\textsubscript{$\pm$0.03} &  \\
DiffuseReg\cite{zhuo2024diffusereg} & 78.89 & 83.62 & 0.51 & 1.74 &  \\
\midrule
FSDR\cite{qin2023fsdiffreg} & 81.06\textsubscript{$\pm$6.03} & 90.39\textsubscript{$\pm$3.78} & 0.25\textsubscript{$\pm$0.12} & 0.32\textsubscript{$\pm$0.04} & 0.02 \\
CMLP\cite{meng2024correlation} & 83.88\textsubscript{$\pm$4.50} & 91.60\textsubscript{$\pm$3.21} & 0.24\textsubscript{$\pm$0.18} & 0.32\textsubscript{$\pm$0.04} & 0.05 \\
Ours (1 step) & 82.94\textsubscript{$\pm$4.98} & 91.27\textsubscript{$\pm$3.35} & 0.30\textsubscript{$\pm$0.19} & 0.36\textsubscript{$\pm$0.06} & 0.06 \\
Ours (2 steps) & 84.27\textsubscript{$\pm$4.10} & 91.86\textsubscript{$\pm$3.05} & 0.38\textsubscript{$\pm$0.25} & 0.39\textsubscript{$\pm$0.11} & 0.19 \\
Ours (10 steps) & \textbf{84.68}\textsubscript{$\pm$3.80} & \textbf{92.15}\textsubscript{$\pm$2.92} & 0.44\textsubscript{$\pm$0.30} & 0.40\textsubscript{$\pm$0.11} & 1.30 \\
\bottomrule
\end{tabular}
\end{table}

%% file: table_dsc.tex
\begin{table}[!b]
\centering
\fontsize{8}{9.6}\selectfont
\caption{Task-wise Dice score comparison (mean$\pm$SD). $\to$MM2 denotes cross-dataset generalization. LV: left ventricle; Myo: myocardium; RV: right ventricle; Mean: average Dice score of LV/Myo/RV; FG: foreground Dice score on merged classes. \textbf{Bold} indicates the higher value within each task and metric. $p$: Bonferroni-corrected paired t-test on Mean Dice score (Ours $>$ CMLP).}
\label{tab:dsc}
\begin{tabular}{llllllll}
\toprule
Dataset & Method & LV & Myo & RV & Mean & FG & $p$ \\
\midrule
\multirow{2}{*}{\makecell[l]{ACDC\\$\to$ED}} & CMLP & 89.59\textsubscript{$\pm$6.24} & 78.66\textsubscript{$\pm$5.22} & 83.39\textsubscript{$\pm$7.01} & 83.88\textsubscript{$\pm$4.50} & 91.60\textsubscript{$\pm$3.21} & \multirow{2}{*}{2e-4} \\
 & Ours & \textbf{90.49}\textsubscript{$\pm$5.02} & \textbf{79.08}\textsubscript{$\pm$4.60} & \textbf{84.46}\textsubscript{$\pm$6.47} & \textbf{84.68}\textsubscript{$\pm$3.80} & \textbf{92.15}\textsubscript{$\pm$2.92} &  \\
\midrule
\multirow{2}{*}{\makecell[l]{ACDC\\$\to$ES}} & CMLP & 80.15\textsubscript{$\pm$12.57} & 79.58\textsubscript{$\pm$5.51} & 76.57\textsubscript{$\pm$8.97} & 78.77\textsubscript{$\pm$6.73} & 88.48\textsubscript{$\pm$4.66} & \multirow{2}{*}{1e-7} \\
 & Ours & \textbf{82.09}\textsubscript{$\pm$11.24} & \textbf{80.70}\textsubscript{$\pm$4.69} & \textbf{76.74}\textsubscript{$\pm$8.96} & \textbf{79.84}\textsubscript{$\pm$6.11} & \textbf{88.60}\textsubscript{$\pm$4.58} &  \\
\midrule
\multirow{2}{*}{\makecell[l]{MM2\\$\to$ED}} & CMLP & 82.01\textsubscript{$\pm$6.69} & 74.17\textsubscript{$\pm$5.01} & 89.79\textsubscript{$\pm$3.67} & 81.99\textsubscript{$\pm$3.87} & 89.18\textsubscript{$\pm$3.53} & \multirow{2}{*}{1e-6} \\
 & Ours & \textbf{83.85}\textsubscript{$\pm$5.61} & \textbf{74.48}\textsubscript{$\pm$4.89} & \textbf{90.23}\textsubscript{$\pm$3.05} & \textbf{82.86}\textsubscript{$\pm$3.31} & \textbf{90.11}\textsubscript{$\pm$2.83} &  \\
\midrule
\multirow{2}{*}{\makecell[l]{MM2\\$\to$ES}} & CMLP & 74.07\textsubscript{$\pm$9.65} & 74.92\textsubscript{$\pm$5.15} & 81.13\textsubscript{$\pm$7.87} & 76.71\textsubscript{$\pm$6.11} & 84.32\textsubscript{$\pm$5.70} & \multirow{2}{*}{4e-3} \\
 & Ours & \textbf{74.64}\textsubscript{$\pm$9.43} & \textbf{75.25}\textsubscript{$\pm$4.99} & \textbf{81.24}\textsubscript{$\pm$7.78} & \textbf{77.04}\textsubscript{$\pm$5.92} & \textbf{84.58}\textsubscript{$\pm$5.66} &  \\
\midrule
\multirow{2}{*}{\makecell[l]{$\to$MM2\\$\to$ED}} & CMLP & 79.82\textsubscript{$\pm$7.71} & \textbf{74.48}\textsubscript{$\pm$5.20} & 88.65\textsubscript{$\pm$3.91} & 80.98\textsubscript{$\pm$4.41} & 87.76\textsubscript{$\pm$4.15} & \multirow{2}{*}{1e-5} \\
 & Ours & \textbf{81.05}\textsubscript{$\pm$7.16} & 74.35\textsubscript{$\pm$5.03} & \textbf{89.04}\textsubscript{$\pm$3.78} & \textbf{81.48}\textsubscript{$\pm$4.04} & \textbf{88.46}\textsubscript{$\pm$3.81} &  \\
\midrule
\multirow{2}{*}{\makecell[l]{$\to$MM2\\$\to$ES}} & CMLP & 73.81\textsubscript{$\pm$9.92} & 75.08\textsubscript{$\pm$5.19} & 80.67\textsubscript{$\pm$8.06} & 76.52\textsubscript{$\pm$6.30} & \textbf{84.35}\textsubscript{$\pm$5.74} & \multirow{2}{*}{0.11} \\
 & Ours & \textbf{73.89}\textsubscript{$\pm$9.88} & \textbf{75.20}\textsubscript{$\pm$5.05} & \textbf{81.11}\textsubscript{$\pm$7.86} & \textbf{76.73}\textsubscript{$\pm$6.13} & 84.14\textsubscript{$\pm$5.91} &  \\
\midrule
\multirow{2}{*}{Avg} & CMLP & 79.91 & 76.15 & 83.36 & 79.81 & 87.61 & \\
 & Ours & \textbf{81.00} & \textbf{76.51} & \textbf{83.80} & \textbf{80.44} & \textbf{88.01} & \\
\bottomrule
\end{tabular}
\end{table}

%% file: table_iterative.tex
\begin{table}[t]
\centering
\fontsize{8}{9.6}\selectfont
\caption{Mean Dice score (\%) under iterative inference and optimisation. Top: performance versus inference steps. Bottom: instance optimisation steps applied to the predicted DDF. \textbf{Bold} indicates the highest value in each column.}
\label{tab:iterative}
\begin{tabular}{c|cccc|lllll}
\toprule
 & & & & & \multicolumn{5}{c}{Inference Steps} \\
Method & SDE & Heun & IG & Guidance & 1 & 2 & 5 & 10 & 20 \\
\midrule
CMLP & & & & & 79.81 & 79.83 & 79.54 & 79.42 & 79.37 \\
\multirow{5}{*}{Ours}
&  & Y & & & 78.92 & 79.76 & 80.12 & 80.24 & 80.30 \\
& Y &  & & & 78.92 & 79.62 & 80.17 & 80.33 & 80.38 \\
& Y & Y & & & 78.92 & 79.76 & 80.28 & 80.38 & 80.37 \\
& Y & Y & Y & & 78.92 & 80.14 & 80.34 & 80.38 & 80.37 \\
& Y & Y & Y & Y & 78.97 & \textbf{80.18} & \textbf{80.39} & \textbf{80.44} & \textbf{80.42} \\
\midrule
& & & & & \multicolumn{5}{c}{Instance Optimisation Steps} \\
Initial DDF & SDE & Heun & IG & Guidance & 1 & 2 & 5 & 10 & 20 \\
\midrule
CMLP & & & & & \textbf{79.82} & 79.83 & 79.85 & 79.87 & 79.87 \\
Ours (1 step) & Y & Y & Y & & 78.94 & 78.95 & 78.98 & 79.00 & 79.02 \\
Ours (2 steps) & Y & Y & Y & & 80.15 & 80.16 & 80.19 & 80.22 & 80.23 \\
\bottomrule
\end{tabular}
\end{table}

%% file: table_clinical.tex
\begin{table}[t]
\centering
\fontsize{8}{9.6}\selectfont
\caption{Mean absolute errors for left/right ventricular ejection fraction (LVEF/RVEF, \%) and myocardial thickness (MT, mm). $\to$MM2 denotes cross-dataset generalization. \textbf{Bold} indicates lower MAE. $p_\text{LV}$/$p_\text{RV}$: Bonferroni-corrected paired $t$-tests on LVEF/RVEF MAE (Ours $<$ CMLP).}
\label{tab:clinical}
\begin{tabular}{lllllll}
\toprule
Dataset & Method & LVEF & RVEF & MT & $p_\text{LV}$ & $p_\text{RV}$ \\
\midrule
\multirow{2}{*}{\makecell[l]{ACDC\\$\to$ED}} & CMLP & 5.06\textsubscript{$\pm$3.78} & 14.83\textsubscript{$\pm$7.79} & 1.01\textsubscript{$\pm$1.18} &  &  \\
 & Ours & \textbf{4.00}\textsubscript{$\pm$2.83} & \textbf{12.01}\textsubscript{$\pm$6.92} & \textbf{0.86}\textsubscript{$\pm$0.96} & 2e-4 & 1e-8 \\
\midrule
\multirow{2}{*}{\makecell[l]{ACDC\\$\to$ES}} & CMLP & 10.98\textsubscript{$\pm$9.02} & 21.41\textsubscript{$\pm$10.61} & 1.60\textsubscript{$\pm$1.98} &  &  \\
 & Ours & \textbf{8.23}\textsubscript{$\pm$7.39} & \textbf{20.31}\textsubscript{$\pm$10.39} & \textbf{1.29}\textsubscript{$\pm$1.73} & 2e-8 & 2e-6 \\
\midrule
\multirow{2}{*}{\makecell[l]{MM2\\$\to$ED}} & CMLP & 13.21\textsubscript{$\pm$7.99} & 4.87\textsubscript{$\pm$3.47} & \textbf{0.62}\textsubscript{$\pm$0.55} &  &  \\
 & Ours & \textbf{8.74}\textsubscript{$\pm$6.71} & \textbf{3.83}\textsubscript{$\pm$3.00} & 0.64\textsubscript{$\pm$0.58} & 2e-20 & 5e-4 \\
\midrule
\multirow{2}{*}{\makecell[l]{MM2\\$\to$ES}} & CMLP & 22.53\textsubscript{$\pm$12.01} & 10.54\textsubscript{$\pm$6.45} & \textbf{0.60}\textsubscript{$\pm$0.49} &  &  \\
 & Ours & \textbf{19.76}\textsubscript{$\pm$11.87} & \textbf{8.37}\textsubscript{$\pm$5.77} & 0.69\textsubscript{$\pm$0.65} & 8e-15 & 2e-16 \\
\midrule
\multirow{2}{*}{\makecell[l]{$\to$MM2\\$\to$ED}} & CMLP & 17.61\textsubscript{$\pm$9.28} & 6.61\textsubscript{$\pm$3.69} & \textbf{0.57}\textsubscript{$\pm$0.48} &  &  \\
 & Ours & \textbf{14.00}\textsubscript{$\pm$8.38} & \textbf{5.33}\textsubscript{$\pm$3.54} & 0.65\textsubscript{$\pm$0.56} & 1e-29 & 5e-19 \\
\midrule
\multirow{2}{*}{\makecell[l]{$\to$MM2\\$\to$ES}} & CMLP & 22.45\textsubscript{$\pm$11.84} & 10.37\textsubscript{$\pm$6.69} & \textbf{0.51}\textsubscript{$\pm$0.38} &  &  \\
 & Ours & \textbf{21.64}\textsubscript{$\pm$12.67} & \textbf{8.87}\textsubscript{$\pm$5.92} & 0.70\textsubscript{$\pm$0.69} & 0.03 & 7e-11 \\
\midrule
\multirow{2}{*}{Avg} & CMLP & 15.31 & 11.44 & 0.82 & & \\
 & Ours & \textbf{12.73} & \textbf{9.79} & \textbf{0.81} & & \\
\bottomrule
\end{tabular}
\end{table}